\documentclass{article}

    \PassOptionsToPackage{numbers}{natbib}

    \usepackage[preprint]{neurips_2022}

\usepackage[utf8]{inputenc} 
\usepackage[T1]{fontenc}    
\usepackage{hyperref}       
\usepackage{url}            
\usepackage{booktabs}       
\usepackage{amsfonts}       
\usepackage{nicefrac}       
\usepackage{microtype}      
\usepackage{xcolor}         

\usepackage{amsmath, amssymb}
\usepackage{bm}
\usepackage[pdftex]{graphicx}
\usepackage{enumitem}
\usepackage{booktabs}
\usepackage{caption}
\usepackage{subcaption}
\usepackage{algorithm, algorithmic}
\usepackage{wrapfig}

\def\E{\mathbb{E}}

\newcommand{\argmin}{\mathop{\mbox{argmin}}}

\usepackage{xcolor}

\newcommand{\mvnnote}[1]{\textcolor{teal}{}}
\newcommand{\knote}[1]{\textcolor{red}{}}
\newcommand{\danielle}[1]{\textcolor{orange}{}}
\newcommand{\hao}[1]{\textcolor{blue}{}}
\newcommand{\huibin}[1]{\textcolor{brown}{}}

\title{Cross-Frequency Time Series Meta-Forecasting}

\author{%
  Mike Van Ness\thanks{Work done while doing an internship at AWS AI Labs.} \\
  Stanford University \\
   \And
   Huibin Shen \thanks{Correspondence to \texttt{huibishe@amazon.com}}\\
   AWS AI Labs \\
  \And
   Hao Wang \\
   AWS AI Labs \\
  \AND
   Xiaoyong Jin \\
   AWS AI Labs \\
  \And
   Danielle C. Maddix \\
   AWS AI Labs \\
  \And
   Karthick Gopalswamy \\
   AWS AI Labs \\
}

\begin{document}

\maketitle

\begin{abstract}
Meta-forecasting is a newly emerging field which combines meta-learning and time series forecasting. The goal of meta-forecasting is to train over a collection of source time series and generalize to new time series one-at-a-time. Previous approaches in meta-forecasting achieve competitive performance, but with the restriction of training a separate model for each sampling frequency. In this work, we investigate meta-forecasting over different sampling frequencies, and introduce a new model, the Continuous Frequency Adapter (CFA), specifically designed to learn frequency-invariant representations. We find that 
CFA greatly improves performance when generalizing to unseen frequencies, providing a first step towards forecasting over larger multi-frequency datasets.\hao{would be nice maybe to have a brief real-world example on adaptation across time series with different frequencies. Currently the abstract reads clear but a bit bland. } \mvnnote{Do you have a suggestion for this? That's one problem with frequency generalization, is it isn't the most useful in practice.} \hao{Just floating some ideas here. For example, different electric appliances may have different cycles (and time duration per cycle) of energy consumption; stock prices from different industries may have different cycles of ups and downs. }
\knote{@MVN: why not speak about its use case in building foundation models that can potentially scale DL for time series? Right now, there is no successful way to scale up datasets in time series forecasting without loosing performance. If CFA can solve the issue with different freq, then one can combine multiple TS in single data to train large models}
\end{abstract}

\section{Introduction}
\vspace{-0.5pc}



Time series forecasting is a classical statistical problem with practical applications in several fields, such as finance, business management \cite{petropoulos2022forecasting}. 
Local statistical models such as ARIMA and ETS \cite{hyndman2018forecasting} have long been state-of-the-art for forecasting.
In recent years, much effort has been put into matching the performance of local models with deep learning approaches, particularly when modeling several closely-related time series \cite{salinas2020deepar, oreshkin2019n, challu2022n}. 

When less data is available in a target dataset, transfer learning from source to target data is often necessary to compete with local methods.
One approach is meta-learning, or domain generalization, where a model is trained to generalize to new target domains after an initial training phase on source data.
Recent work has shown that meta-learning for time series forecasting, or meta-forecasting, can achieve competitive performance with only local fine-tuning \cite{grazzi2021meta} or even with no fine-tuning \cite{oreshkin2021meta}.  
Such approaches are \textit{zero-shot} forecasters, as they can forecast out-of-domain time series one-at-a-time without access to any related time series.

\begin{figure}
    \centering
    \includegraphics[width=0.8\linewidth]{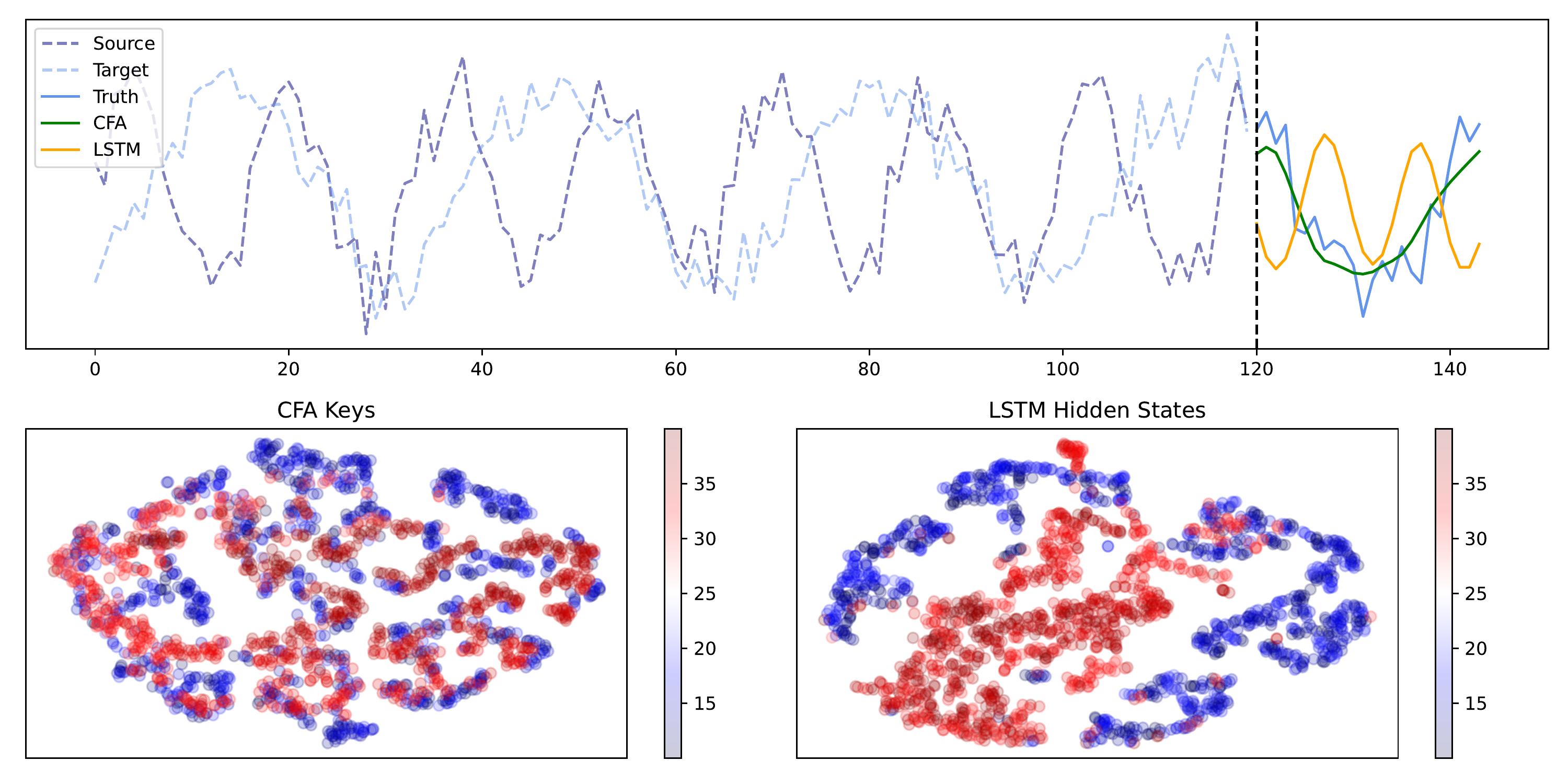}
    \caption{For frequency generalization, CFA outperforms LSTM. Both models are trained on synthetic source data with period length randomly sampled between [10, 20], and are zero-shot applied to synthetic target data with period length randomly sampled from [30, 40]. In the top plot, the LSTM model predicts the wrong seasonality, whereas CFA successfully adapts to the new seasonality. In the bottom plot, we see that CFA keys are invariant to frequency (color), but LSTM hidden states have different distributions for source and target frequencies ranges.}
    \label{fig:intro_fig}
    \vspace{-1pc}
\end{figure}

Almost all of the previous transfer learning works use the assumption that source and target data come from the same sampling frequency, e.g. hourly, daily, monthly, etc. 
We propose a different assumption: \textit{all data is seasonal, but not necessarily from the same sampling frequency}. The typical seasonality associated with each sampling frequency then creates a correspondence between sampling frequency and signal frequency. As seen in Figure \ref{fig:intro_fig}, seasonal time series of different signal frequencies appear very similar to humans, but are challenging for typical forecasting models to transfer between.
Along with our data assumption, we consider a new task, \textit{frequency generalization}, in which we task a model to generalize to \textit{unseen} frequencies during meta-test time. For successful frequency generalization, we propose a new model, the Continuous Frequency Adapter (CFA). As shown in Figure \ref{fig:intro_fig}, CFA can forecast the correct frequency on data of new unseen frequencies, which other methods cannot.
CFA uses continuous domain adaptation \cite{wang2020continuously} to enforce frequency-invariant hidden states, which is vital for frequency generalization. We summarize the novelty and contribution of this paper:
\begin{itemize}[itemsep=0ex, topsep=0ex, partopsep=0ex, parsep=0.5ex,
		leftmargin=2.1ex]
	\item We explore meta-learning over different sampling frequencies, which is previously unexplored. For this, we introduce a new task, \textit{frequency generalization}, in which we task a model to generalize to time series of new sampling frequencies during test time.
	\item We develop a new model, CFA, which achieves improved performance for frequency generalization than previous meta-learning models. CFA uses continuous domain adaptation \cite{wang2020continuously} to adapt to new sampling frequencies, a new technique for the time series literature. Specifically, CFA uses signal frequencies obtained from a Fourier transform of the input time series to define continuous domain indices, a novel technique for time series domain generalization.
\end{itemize}

\paragraph{Related Work}


Transfer learning has been explored previously for time series through several subfields.
Time series representation learning \cite{yue2022ts2vec, zerveas2021transformer} does self-supervised pretraining for time series data. Domain adaptation approaches \cite{jin2022domain, hu2020datsing} learn models that can adapt from one source dataset to a different target dataset, often utilizing adversarial training.  In few-shot learning \cite{iwata2020few}, a model transfers knowledge by learning how to learn from a small support set of related time series. 

Recently, a few papers have particularly addressed meta-forecasting.
In \cite{oreshkin2021meta}, the popular forecasting model N-BEATS is shown to fit a meta-learning framework, and achieves competitive zero-shot performance. 
In Meta-GLAR \cite{grazzi2021meta}, a local closed-form head is used to adapt global representations to new time series. Our work is most similar to these meta-learning approaches in that performance on the target dataset is evaluated in a zero-shot manner. Our model, however, takes inspiration primarily from \cite{jin2022domain} in its use of adversarial training and self-attention. We emphasize that all of the above cited papers only consider transferring between datasets of the same frequency, with the exception of \cite{jin2022domain} which considers domain adaptation opposed to the harder task of zero-shot meta-learning.

\section{Problem Definition}\label{sec:prob_def}
\vspace{-0.5pc}

Time series forecasting is the problem of predicting future observations, i.e. forecasting, given a past context window. That is, for some time series $(\bm{z}_t)_{t > 0}$, data samples are of the form
\[
\bm{x} = \bm{z}_{1:\tau_c}, \quad \bm{y} = \bm{z}_{\tau_c+1 : \tau_c + \tau_f}
\]
where $\tau_c$ is the length of the context and $\tau_f$ is the length of the forecast. A model $f$ then estimates $\bm{y}$ from $\bm{x}$. 
If $f$ has parameters $\theta$, we aim to find the parameters $\theta$ that minimize the forecasting loss, i.e.
\begin{equation}
\label{eq:loss_fn}
    \argmin_\theta \E[L_f(F(\bm{x}), \bm{y}; \theta)] 
\end{equation}
where $L_f$ is a forecasting loss such as MSE. 

What distribution the expectation is taken over, i.e. what distribution $\bm{z}$ comes from, depends on the nature of the forecasting problem.
In this paper, we consider the \textit{zero-shot} framework, in which case $\bm{z} \sim \mathcal{T}$ comes from some target distribution $\mathcal{T}$, but we only have training data $(\bm{x}_1, \bm{y}_1), \ldots, (\bm{x}_n, \bm{y}_n) \sim \mathcal{S}$ from a source distribution $\mathcal{S}$.
This defines the problem of \textit{meta-forecasting}, where the model $f$ must meta-learn on $\mathcal{S}$ with the goal of minimizing \ref{eq:loss_fn} on $\mathcal{T}$ with no additional training on $\mathcal{T}$, where $\mathcal{T}$ could be \textit{any} target distribution unseen during training.

In this paper, we focus on \textit{frequency generalization}, which we define as meta-forecasting with $\mathcal{S}$ and $\mathcal{T}$ representing distinct frequencies. 
This is more challenging than the setting considered in previous meta-forecasting papers, in which $\mathcal{T}$ only contains frequencies that are already seen in $\mathcal{S}$. 
We do explore this second easier setting to compare to previous papers, and present the results in Appendix \ref{subsec:uni_vs_multi}.

Since minimizing \ref{eq:loss_fn} on all potential $\mathcal{T}$ is an ambitious and likely unrealistic desire, we restrict $\mathcal{S}$ and $\mathcal{T}$ to be distributions representing seasonal datasets.
Under this assumption, different sampling frequencies correspond to different signal frequencies, and thus generalizing to new sampling frequencies corresponds to generalizing to new signal frequencies.
This makes the frequency generalization problem realistic, since seasonal data of different signal frequencies appear quite similar to the human eye but are difficult for machine learning models to generalize between as shown in Figure \ref{fig:intro_fig}.
Relaxing this assumption would be significantly more challenging, and we leave this to future work.



\section{Continuous Frequency Adaptation}\label{sec:cfa}
\vspace{-0.5pc}

Most meta-forecasting models struggle to learn frequency-invariant signal vital for meta-forecasting. This challenge is demonstrated in the bottom right portion of Figure \ref{fig:intro_fig}, where an LSTM model learns hidden states whose distribution depends on the input signal frequency.

To overcome this challenge, we introduce the Continuous Frequency Adapter (CFA), which is specifically designed to learn frequency-invariant representations (see bottom left of Figure \ref{fig:intro_fig}). 
CFA is primarily a self-attention network, and utilizes adversarial training to enforce frequency invariance in the attention keys and queries (but not the values). The model architecture is inspired by the Domain Adaptation Forecaster (DAF) \cite{jin2022domain}, but uniquely 
uses continuous domain indices \cite{wang2020continuously} obtained by a Fourier transform to generalize to unseen signal frequencies.
\begin{figure}
    \centering
    \begin{minipage}[c]{0.57\textwidth}
    \includegraphics[width=\linewidth]{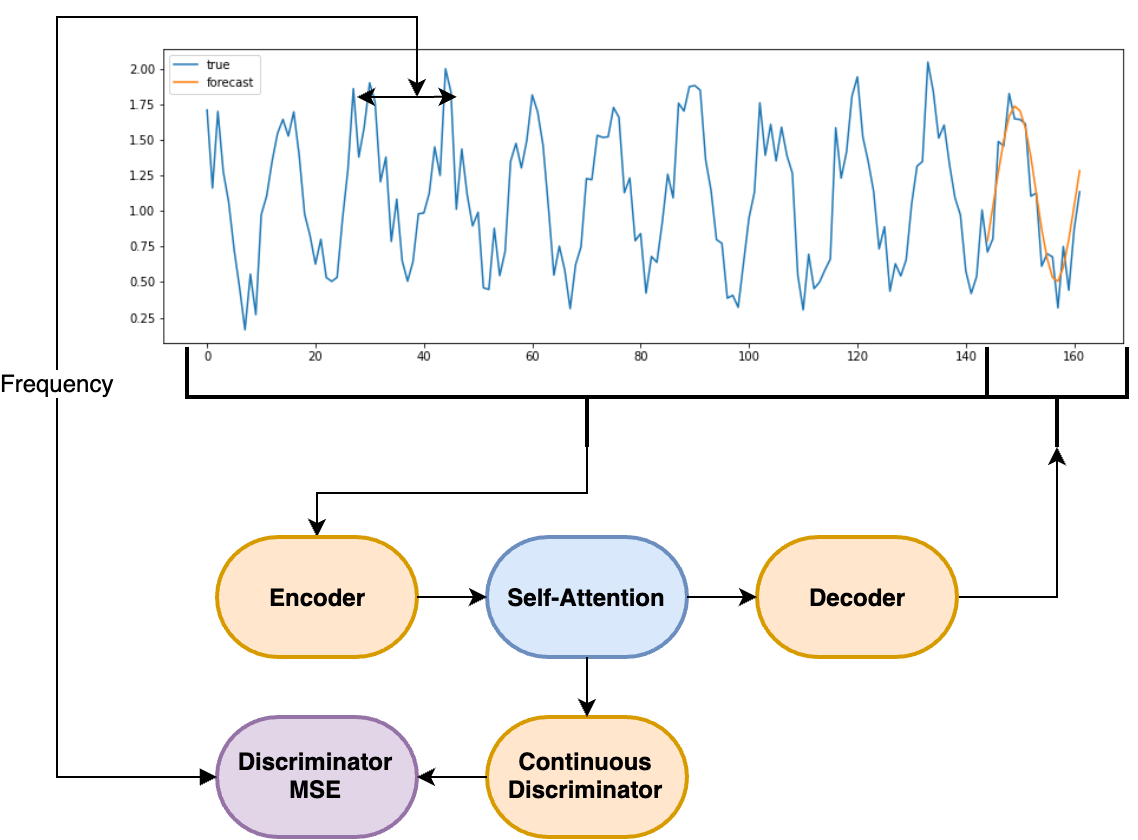}
    \end{minipage}
    \hfill
    \begin{minipage}[c]{0.4\textwidth}
    \caption{CFA Architecture. The encoder takes in a context window and produces keys, queries, and values, which are used by the self-attention module to produce representations for each timestep in the forecast window, which the decoder uses to make forecasts. Meanwhile, the keys and queries are passed to the continuous discriminator, which predicts the top $k$ frequencies of the context window sorted by FFT amplitude. Adversarial training via Equation \ref{eq:minimax_loss} is used to learn good forecasts while also making keys and queries frequency-invariant.}
    \label{fig:cfa_architecture}
    \end{minipage}
    \vskip -0.5cm
\end{figure}

\paragraph{CFA Architecture}

The CFA architecture is summarized in Figure \ref{fig:cfa_architecture}.
The encoder, self-attention, and decoder blocks are similar to the DAF architecture \cite{jin2022domain}.
The encoder consists of a position-wise MLP follows by a series of 1D convolutional layers, and the decoder is a position-wise MLP.
The self-attention block is a standard multi-head attention block as in transformer architectures \cite{vaswani2017attention}. 
The CFA discriminator is an MLP, like in DAF, but unlike DAF, our discriminator outputs a continuous response to match the continuous domain index (see next paragraph).
Also, unlike DAF, the encoder and decoder blocks are shared for all time series, since a continuous domain index does not allow for separate encoders and decoders for each possible index since there are infinitely many domain indices.

The discriminator takes as input all keys and queries from the self-attention block, in order to enforce domain invariance in the keys and queries via adversarial training (see Equation \ref{eq:minimax_loss}). 
The motivation for this choice is that the keys and queries are used to generate the attention weights, which tell the model, for any given time point, which other time points are most relevant.
For time series data, especially seasonal time series data, this importance weighting can be independent of the signal frequency, as the attention weights only need to capture the phase of the signal.
Meanwhile, the values in the self-attention block are left independent of the discriminator, and thus can learn information that is dependent on the given time series, e.g. what the time series typically looks like at each phase. 

\paragraph{Continuous Domain Generalization}

A key component of CFA is the use of continuous domain indices \cite{wang2020continuously}. 
These continuous domain indices serve as labels for the discriminator, which takes as input the self-attention keys and queries and outputs a continuous domain index prediction.
The domain indices are obtained via an FFT on the inputted context window to capture the signal frequencies of the time series.
Specifically, we take the absolute value of the FFT outputs to obtain the amplitude corresponding to each frequency bin. 
We then select the top $k$ frequencies, sorted by their amplitudes, and use their inverses (i.e. the period lengths) as the discriminator labels. 
We normalize the labels to be between 0 and 1 to stabilize the discriminator loss.
For synthetic data we use $k=1$, since the synthetic data is noist sine waves without subfrequencies, and on real data we use $k=2$.

\paragraph{Adversarial Loss}

CFA utilizes adversarial training via a typical minimax loss \cite{goodfellow2020generative}. Let $E$ be an encoder, $F$ a forecasting decoder, and $D$ a discriminator (for CFA, E generates the self-attention keys, queries, and values, and F produces forecasts using these self-attention inputs). The goal of CFA is to solve the following minimax problem: 
\begin{equation} \label{eq:minimax_loss}
\min_{E, F} \max_D \E[L_f(\bm{x}; E, F)] - \lambda \E[L_d(\bm{x}; E, D)]
\end{equation}
where $L_f$ is the forecasting loss and $L_d$ is the discriminator loss. 
In words, $D$ is trained to minimize $L_d$, thereby training a strong discriminator, while $E$ and $F$ are trained to both produce good forecasts (i.e. minimize $L_f$) while maintaining hidden states that the strong discriminator cannot predict well (i.e. maximizing $L_d$). 
Such adversarial training allows the model to learn frequency-invariant discriminator inputs (keys and queries) while still producing good forecasts.   
In practice, the forecast/generative parameters (E, F) and the discriminative parameters (D) are updated in an alternating fashion, see Algorithm \ref{alg:cfa_alg}.

\paragraph{Training Procedure}

Since CFA is designed to learn frequency-invariant keys and queries, it is essential that CFA is trained over source data with varied frequencies.
The multi-source training procedure is illustrated in Algorithm \ref{alg:cfa_alg}.
The training works by sampling one batch from each of the source datasets, and updating the generative and discriminative parameters from the sum of the batch losses. 
The forecast/generative parameters (E, F) and the discriminative parameters (D) are updated in an alternating fashion, as typical for adversarial training.

\begin{algorithm}[h!]
\caption{CFA Training Algorithm}
\begin{algorithmic}[1]\label{alg:cfa_alg}
    \STATE \textbf{Input:} source datasets $S_1, \ldots S_d$, forecast loss $L_f$, discriminator loss $L_d$.
    \FOR{epoch $=1$ {\bfseries to} $E$}
        \FOR{$i=1$ {\bfseries to} n\_batches\_per\_epoch}
            \STATE Sample $\bm{x}_j, \bm{y}_j \sim S_j$ for $j=1,\ldots,d$
            \STATE Compute generative loss $L_j = L_f(\bm{x}_j, \bm{y}_j) - \lambda L_d(\bm{x}_j)$ for each $j$
            \STATE Update generative parameters via $L = L_1 + \cdots + L_d$
            \STATE Compute discriminative loss $L_j = L_d(\bm{x}_j, \text{FFT}(\bm{x}_j)$ for each $j$
            \STATE Update discriminative parameters via $L = L_1 + \cdots + L_d$
        \ENDFOR
    \ENDFOR
\end{algorithmic}
\end{algorithm}

\section{Experiments}
\vspace{-0.5pc}

\paragraph{Models} For our experiments, we consider the following models. 
\begin{itemize}[itemsep=0ex, topsep=0ex, partopsep=0ex, parsep=0.5ex,
		leftmargin=2.1ex]
    \item \textbf{Mean}: simple baseline that forecasts the mean from the context window.
    \item \textbf{LSTM}: an auto-regressive LSTM model similar to DeepAR \cite{salinas2020deepar}.
    \item \textbf{NBEATS}: deep model with mostly linear layers and doubly-residual connections \cite{oreshkin2019n}.
    \item \textbf{CFA}: Our model described in Section \ref{sec:cfa}.
\end{itemize}

Since frequency generalization is a new task, there are some baselines which cannot readily be adapted.
For one, any method which requires separate modules for different sampling frequencies cannot be used, e.g. DAF \cite{jin2022domain}, because it is impossible to train a new module for the target sampling frequency in the zero-shot regime.
Additionally, we think that is it critical on real data to have different forecasting lengths for different sampling frequencies, and thus require models that can forecast an arbitrary number of time steps ahead during test time.
CFA and LSTM can easily do this since they are autoregressive forecasters, i.e. they use the previous forecasts to make each successive forecast. 
NBEATS, on the other hand, requires a fixed forecast length that cannot be adjusted during test time, and thus we do not use it as a baseline for real data experiments.
We still consider NBEATS as a baseline for synthetic data generated with uniform forecast length, though, since NBEATS has been shown to be a strong meta-forecaster \cite{oreshkin2021meta}. 

\paragraph{Synthetic Data}

We generate synthetic time series data using sinusoidal curves with Gaussian noise and uniformly random period length (inverse of frequency), see Appendix \ref{subsec:dataset_details} for full details. 
We designate one period range for source data and one period range for target data. Models are trained on source data and applied zero-shot to target data, evaluated by mean squared error (MSE) in the forecast window. 
Results are shown in Table \ref{tab:synthetic_zero_shot}.
Across all combinations of source and target period ranges, CFA is either the best model or within one standard deviation of the best model. As shown in Figure \ref{fig:intro_fig}, CFA is able to learn good forecasts on the source data while maintaining frequency-invariant keys and queries, allowing CFA to generalize to new frequencies. In comparison, LSTM and NBEATS learn frequency-dependent signal, and thus fail to generalize to new frequencies, even failing to beat the simple mean baseline.

\begin{table}
    \centering
    \scalebox{0.8}{
    \begin{tabular}{llllll}
    \toprule
    Source Range & Target Range & Mean & CFA & LSTM & NBEATS \\
    \midrule
    (10, 15) & (15, 20) &  $0.260 \pm 0.002$ & $\bm{0.088 \pm 0.011}$ &  $0.312 \pm 0.057$ &  $0.424 \pm 0.017$ \\
             & (20, 25) &  $0.259 \pm 0.003$ & $\bm{0.190 \pm 0.025}$ &  $0.456 \pm 0.081$ &  $0.351 \pm 0.007$ \\
             & (25, 30) &  $\bm{0.262 \pm 0.003}$ & $\bm{0.223 \pm 0.041}$ &  $0.426 \pm 0.028$ &   $0.320 \pm 0.004$ \\
    (15, 20) & (10, 15) &  $0.259 \pm 0.003$ & $\bm{0.071 \pm 0.014}$ &  $0.388 \pm 0.035$ &  $0.417 \pm 0.011$ \\
             & (20, 25) &  $0.259 \pm 0.003$ & $\bm{0.055 \pm 0.004}$ &  $0.185 \pm 0.055$ &  $0.469 \pm 0.011$ \\
             & (25, 30) &  $0.262 \pm 0.003$ & $\bm{0.082 \pm 0.007}$ &  $0.499 \pm 0.072$ &  $0.526 \pm 0.012$ \\
    (20, 25) & (10, 15) &  $\bm{0.259 \pm 0.003}$ & $\bm{0.209 \pm 0.099}$ &  $0.497 \pm 0.037$ &  $0.333 \pm 0.005$ \\
             & (15, 20) &  $0.260 \pm 0.002$ & $\bm{0.066 \pm 0.013}$ &  $0.282 \pm 0.082$ &  $0.425 \pm 0.008$ \\
             & (25, 30) &  $0.262 \pm 0.003$ & $\bm{0.064 \pm 0.006}$ &  $0.134 \pm 0.052$ &   $0.387 \pm 0.020$ \\
    (25, 30) & (10, 15) &  $\bm{0.259 \pm 0.003}$ & $\bm{0.257 \pm 0.056}$ &  $0.442 \pm 0.015$ &  $\bm{0.303 \pm 0.006}$ \\
             & (15, 20) &  $\bm{0.260 \pm 0.002}$ & $\bm{0.213 \pm 0.081}$ &  $0.533 \pm 0.045$ &  $0.417 \pm 0.007$ \\
             & (20, 25) &  $0.259 \pm 0.003$ & $\bm{0.069 \pm 0.015}$ &  $0.177 \pm 0.027$ &  $0.397 \pm 0.021$ \\
    \bottomrule
    \end{tabular}
    }
    \vspace{0.5pc}
    \caption{Frequency generalization on synthetic data, measured as MSE of forecast. Source and target range indicate the range of uniformly random period lengths in the source and target data, respectively. For each source/target pair, the model is trained on source and applied zero-shot to target. Across all pairs of ranges, CFA has the best performance. }
    \label{tab:synthetic_zero_shot}
    \vspace{-2pc}
\end{table}

\paragraph{Real Data}

We focus on real world datasets that exhibit clear seasonality. 
We use the following datasets, each with a different sampling frequency: elec (hourly), uber (daily), tourism monthly (monthly), and tourism quarterly (quarterly).
We load all datasets using GluonTS \cite{alexandrov2020gluonts}. More information on each dataset, data preprocessing, and setup can be found in Appendix \ref{subsec:dataset_details}.
For each experiment, we designate one dataset as the target dataset and use the other 3 as the source datasets. 
We evaluate models by their Normalized Deviation (ND) \cite{jin2022domain} on the forecasting window.
We do not run NBEATS because it requires equal context/forecast lengths across datasets, which we do not enforce for frequency generalization.
The results are shown in Table \ref{tab:freq_gen}. 
As was the case with synthetic data, CFA outperforms LSTM for frequency generalization.
\begin{table}
    \centering
    \begin{minipage}[c]{0.50\textwidth}
    \scalebox{0.64}{
    \begin{tabular}{llll}
    \toprule
    {} & Mean & CFA & LSTM \\
    \midrule
    elec              &  $0.404 \pm 0.000$ & $\bm{0.281 \pm 0.065}$ &  $0.376 \pm 0.031$ \\
    tourism\_monthly   &  $0.312 \pm 0.000$ & $\bm{0.226 \pm 0.019}$ &  $0.296 \pm 0.033$ \\
    tourism\_quarterly &  $0.229 \pm 0.000$ & $\bm{0.178 \pm 0.021}$ &   $\bm{0.19 \pm 0.018}$ \\
    uber              &  $0.203 \pm 0.000$ & $\bm{0.166 \pm 0.012}$ &   $0.252 \pm 0.02$ \\
    \bottomrule
    \end{tabular}
    }
    \end{minipage}
    \begin{minipage}[c]{0.48\textwidth}
    \caption{Frequency generalization on real data, evaluated by test Normaized Deviation. Each dataset has a different frequency, and each row corresponds to one target dataset, using all other datasets as source. As on synthetic data, CFA is better at frequency generalization than LSTM. We do not evaluate N-BEATS in this setting because each dataset has a different context and forecast length, making N-BEATS incompatible.}
    \label{tab:freq_gen}
    \end{minipage}
\end{table}


\section{Conclusion}
\vspace{-0.5pc}

Previous meta-forecasting papers have shown strong performance, but only when training one model per sampling frequency. 
In this paper, we instead consider frequency generalization, i.e. generalizing to unseen frequencies, for which it is not possible to train one model per sampling frequency. 
While previous meta-forecasting models are not successful, our CFA model provides much improved performance. 
This is an important first step towards building forecasting models robust to signal frequency, which could be trained over larger and less constrained datasets.

\bibliographystyle{abbrv}
\bibliography{bib}

\appendix
\section{Appendix}

\subsection{Additional Dataset Details}
\label{subsec:dataset_details}

\paragraph{Synthetic Data}

Synthetic data is generated using GluonTS \cite{alexandrov2020gluonts}. Each time series is generated as the sum of a sine curve and Gaussian noise. The parameter are chosen according to
\begin{itemize}[itemsep=0ex, topsep=0ex, partopsep=0ex, parsep=0.5ex,
		leftmargin=2.1ex]
    \item Phase: $\sim \text{Unif}(0, 2\pi)$.
    \item Period: $\sim \text{Unif}(p_{\min}, p_{\max})$.
    \item Amplitude: $\sim \text{Unif}(0.5, 2)$.
    \item Noise: $\sim \text{Normal}(\mu=0, \sigma=0.2)$.
\end{itemize}
The period range $(p_{\min}, p_{\max})$ for source and target data is indicated in each row of Table \ref{tab:synthetic_zero_shot}. Each time series has 120 samples in the context window, and 24 samples in the forecast window. For each synthetic dataset, we generate 5000 time series with exactly one context length plus one forecast length's worth of data. We designate 4000 time series as the training set and the 1000 as the test set.

\paragraph{Real Data}
Table \ref{tab:real_datasets} shows the details of the datasets used in the real data experiments. All datasets are accessed via GluonTS \cite{alexandrov2020gluonts}, with the GluonTS name for each dataset provided. Following our assumptions, we focus on datasets that exhibit clear seasonality. The hourly, daily, monthly, and quarterly datasets have period lengths of 24, 7, 12, and 4 respectively. 

Unlike on synthetic data, where training and test samples come from different time series, training and test sets come from different time windows across all time series. The last forecast length of each time series is always used as the test set. For the training set, we sample batches using an expected number sampler, such that we sample as many unique context/forecast windows as possible from the training window.

Lastly, on real datasets, it is essential to scale each time series, as is typical in deep learning forecasting approaches. We scale each sample by subtracting by the mean and dividing by the standard deviation from the context window. We then rescale our forecasts and compute the forecast loss with the true unscaled forecast labels. Note that we use Normalized Deviation as our forecasting loss on real data, which itself does normalization and thus should be fed unscaled values.

\begin{table}[h!]
    \centering
    \scalebox{0.8}{
    \begin{tabular}{c|c|c|c|c}
    \hline
    Dataset & GluonTS Name & Frequency & Number of Time Series & Original Source \\
    \hline
    elec & electricity\_nips & hourly & 370 & UCI \cite{Dua:2019} \\
    traffic & traffic\_nips & hourly & 963 & UCI \cite{Dua:2019} \\
    solar & solar\_nips & hourly & 137 & \cite{lai2018modeling} \\
    uber & uber\_tlc\_daily & daily & 262 & fivethirtyeight \cite{fivethirtyeight2015} \\
    NN5 & nn5\_daily\_without\_missing & daily & 111 & NN5 challenge \cite{nn52008} \\
    tourism monthly & tourism\_monthly & monthly & 366 & Tourism competition \cite{athanasopoulos2011tourism} \\
    tourism quarterly & tourism\_quarterly & quarterly & 366 & Tourism competition \cite{athanasopoulos2011tourism} \\
    \hline
    \end{tabular}
    }
    \vspace{0.5pc}
    \caption{Description of all real-world datasets used in experiments.}
    \label{tab:real_datasets}
\end{table}

\subsection{Uni vs Multi Frequency Experiments}\label{subsec:uni_vs_multi}

Previous meta-forecasting papers \cite{grazzi2021meta, oreshkin2021meta} only consider meta-learning across time series of the same sampling frequency. For most of the paper, we focus on frequency generalization, i.e. generalization to unseen frequencies in the target data. Another valid question to ask is: does using source data of multiple frequencies improve target performance even when the target data frequency is already seen? Even if the performance is comparable, multi-frequency source data has the additional benefit of needing to only train one model instead of one model per frequency. 

To investigate this question, we run the following experiment. We designate the following datasets as source datasets: Elec, Uber, Tourism Monthly, Tourism Quarterly, and the following datasets as target datasets: NN5, Solar, Traffic. 
For each target dataset, the multi-source models (Multi-LSTM, Multi-NBEATS, and CFA) are trained jointly on all source datasets, while uni-source models (LSTM, NBEATS) are trained only on the 1 source dataset whose sampling frequency matches the given target dataset.
For Multi-NBEATS, in order to train the model over datasets of different context/forecast lengths, we add an encoder to the inputs and a decoder to the outputs, unique to each frequency, to enforce an equal latent backcast/forecast length for NBEATS.
We evaluate each model by the zero-shot performance on target after training on source, using Normalized Deviation (ND) \cite{jin2022domain} as the evaluation metric.

The results are shown in Table \ref{tab:uni_vs_multi}. We find that the uni-source models always outperform their multi-source counterparts, and CFA has the worst performance for 2 out of 3 target datasets. 
This supports the conclusion that multi-source training deteriorates performance for zero-shot meta-forecasting. 
Further, learning frequency-invariant signal, as CFA does, is not suitable when source data is available of the same frequency as the target data.
This is an area of potential improvement for future work.

\begin{table}
\setlength{\tabcolsep}{5.0pt}
    \centering
    \begin{tabular}{llllll}
    \toprule
    {} &       Multi LSTM &             LSTM &     Multi NBEATS &           NBEATS &              CFA \\
    \midrule
    NN5     &  $0.165 \pm 0.009$ &  $\bm{0.158 \pm 0.006}$ &   $0.22 \pm 0.011$ &   $0.18 \pm 0.005$ &   $0.193 \pm 0.01$ \\
    Solar   &  $0.969 \pm 0.115$ &  $0.891 \pm 0.111$ &  $0.686 \pm 0.021$ &  $\bm{0.602 \pm 0.019}$ &  $1.005 \pm 0.037$ \\
    Traffic &  $0.325 \pm 0.014$ &  $0.298 \pm 0.007$ &  $0.249 \pm 0.003$ &  $\bm{0.238 \pm 0.003}$ &   $0.449 \pm 0.03$ \\
    \bottomrule
    \end{tabular}
    \vspace{0.5pc}
    \caption{Comparison of zero-shot performance for Question 1 on real data. LSTM and NBEATS are trained on one source dataset of the same frequency as the listed target (elec $\to$ solar, traffic, uber $\to$ NN5), while Multi LSTM, Multi NBEATS, and CFA are all trained on 4 source datasets of different frequency (elec, uber, tourism monthly, tourism quarterly). We see that in general, the models trained over uni-frequency source data are superior.}
    \label{tab:uni_vs_multi}
\end{table}

\end{document}